# Principal Phrase Mining

Ellie Small, PhD, (corresponding author): esmall1@drew.edu

Department of Mathematics and Computer Science, Drew University, 36 Madison Ave,

Madison, NJ 07940

Javier Cabrera, PhD

Department of Statistics & Department of Medicine, Rutgers University, New Brunswick, NJ

# ABSTRACT

Extracting frequent words from a collection of texts is commonly performed in many subjects. However, as useful as it is to obtain a collection of commonly occurring words from texts, there is a need for more specific information to be obtained from texts in the form of most commonly occurring *phrases*. Despite this need, extracting frequent phrases is not commonly done due to inherent complications, the most significant being double-counting. Double-counting occurs when words or phrases are counted when they appear inside longer phrases that themselves are also counted, resulting in a selection of mostly meaningless phrases that are frequent only because they occur inside frequent super phrases. Several papers have been written on phrase mining that describe solutions to this issue; however, they either require a list of so-called quality phrases to be available to the extracting process, or they require human interaction to identify those quality phrases during the process.

We present here a method that eliminates double-counting via a unique rectification process that does not require lists of quality phrases. In the context of a set of texts, we define a principal phrase as a phrase that does not cross punctuation marks, does not start with a stop word, with the exception of the stop words "not" and "no", does not end with a stop word, is frequent within those texts without being double counted, and is meaningful to the user. Our method identifies such principal phrases independently without human input, and enables their extraction from any texts within a reasonable amount of time.

# 1. Introduction

A *document* is considered a text together with additional information called meta data, such as the name of the text, its author, the date it was created, etc.

A *corpus* (plural *corpora*) is a collection of documents that are related together with additional data called corpus-level meta data, such as the name of the collection, the relationship between the documents, the date they were combined, etc.

In text mining we wish to investigate and analyze the information provided in texts without actually having to read those texts. In particular, we may wish to obtain relevant information for the documents in a corpus, so we can make better choices as to which document(s) we may wish to investigate further and for which we may actually want to read its text.

With the currently available methods, we can select the most frequent words in a corpus and the number of times they appear in the text of each of its documents. We can do this, for example, using the R text mining package `tm` (Feinerer, Introduction to the tm Package. Text Mining in R, 2017) (Feinerer, An Introduction to Text Mining in R, 2008) (Feinerer, Hornik, & Meyer, Text Mining Infrastructure in R, 2008). This information can indeed be used to provide information about the context of each text without having to read the full text.

However, for many problems it would be more useful to be able to select common *phrases* with their frequencies instead, as they can provide more specific information. But extracting phrases from texts brings with it certain complications that do not occur when we extract words from texts; the most significant of these complications being the issue of double-counting.

*Double-counting* occurs when words or phrases are counted when they appear inside longer phrases that are also counted. For example, when the phrase "severe cardiovascular disease" results in one count of the phrase "severe cardiovascular disease" as well as one count of "severe cardiovascular" and one count of "cardiovascular disease".

A *stop word* is a trivial word such as "a", "and", "of", "the", "we", i.e., a word that does not provide much useful and/or important information. Note that (Fox, 1989) discussed a stop list, which is essentially a list of stop words.

A *principal phrase* is a phrase that occurs in a text or a set of texts, does not cross punctuation marks, does not start with a stop word - with the exception of the stop words "not" and "no" - does not end with a stop word, is frequent within those texts without having been double counted, and is meaningful to the user. Here we present a technique for extracting principal phrases from texts.

## 1.1. Phrase Mining Complications

When we extract words from texts, we usually prepare the texts first. Some of those preparations involve converting everything to lower case, remove punctuation, and remove stop words.

An *n-gram* is any consecutive set of n words, as it appears in a text.

It is not difficult to expand the extraction from words to n-grams. However, this would result in multiple problems, most described in detail by (Liu, Shang, & Han, 2017). Here are the main issues:

- If we continue to exclude stop words, then for example, the phrase "cats and dogs" would turn into the meaningless phrase "cats dogs". But if we include stop words, it could result in other meaningless phrases such as "and this disease" and "attack of". Note also that the words "not" and "no" are generally considered stop words, but ones that will change the meaning of a word or phrase significantly if they occur right before it.
- Some phrases are inherently meaningless to researchers such as the phrase "here we are", which may still occur frequently. In addition, many meaningless phrases are subject dependent. For example, if we obtained a set of texts related to the omicron variant of the coronavirus, the phrase "omicron variant" itself may not be meaningful to the investigator since it will appear in all texts.

- If we remove punctuation marks before extracting phrases, we would be selecting n-grams that cross those punctuation marks, resulting in meaningless phrases.
- Most importantly, we would have double-counting issues.

Note that if we were to allow double-counting, then one word or group of words may be counted in multiple phrases. This would result in meaningless subsets of phrases being presented as frequent phrases together with their meaningful super phrases since they would occur with the same frequency. If we were to cluster the results this would likely cause these phrases to be clustered together, and the meaningless sub phrase would provide no useful additional information to the cluster that wasn't already provided by its super phrase. As an example, consider the phrases "severe cardiovascular disease" and "severe cardiovascular", the former meaningful, the latter meaningless. Since the latter always occurs inside the former, its frequency would be the same, and even if the phrase "severe cardiovascular" never occurred on its own, it would still be presented as a phrase as frequent as "severe cardiovascular disease". If we were to display the most common phrases in a corpus, the phrase "severe cardiovascular disease" would appear right next to the phrase "severe cardiovascular" since its frequency would be the same. In addition, some meaningful phrases such as "cardiovascular disease" would have their frequency exaggerated since the frequencies of "severe cardiovascular disease" would be added to its stand-alone occurrences. This would cause any sub phrase such as "cardiovascular disease" to appear to always be more common than its super phrase.

There are several papers that suggest solutions to these problems. However, most of the current methods for extracting phrases cannot be fully automated as they require human experts to design rules or label phrases.

(Liu, Shang, & Han, 2017) suggest a technique called phrasal segmentation which requires a set of quality phrases to be supplied before the phrases can be extracted. Their method needs domain experts to first carefully select hundreds of quality phrases from millions of candidates. (Shang, et al., 2018) discuss using a method called AutoPhrase, which needs a general knowledge base (e.g., Wikipedia) in the required language, and benefits from a Part-Of-Speech tagger. This method, like many other methods, also seeks to keep a list of so-called quality phrases, which it then compares to the contents of a collection of texts. The key difference is it obtains this list from a general knowledge base, rather than domain experts or the investigators themselves.

We developed a method that avoids the issues described, is simple, requires neither human effort to label phrases nor a general knowledge base, is highly efficient, and extracts principal phrases only. Phrases of any range of number of words can be extracted, including those consisting of just one word. It is able to do this without human input, and as such can easily be automated.

Our method is based on the idea that phrases consisting of many words that are frequent are likely to be important phrases, and are given priority over phrases consisting of fewer words.

## 2. Method

The method we propose starts with a collection of texts, or a corpus of documents containing texts.

We split each text in blocks, which are determined by the placement of punctuation marks in the texts.

For a specific range of n, we extract all n-grams from each block, but exclude those that start with a stop word not equal to either "not" or "no", those that end in a stop word, and those that are in a list of phrases that we wish to exclude.

For each n-gram that makes the cut, we store its location, i.e., the text, block, and position within the block. These are the n-gram locations.

Once all appropriate n-gram locations have been extracted, all n-grams that have n-gram locations go through a rectification process to remove all double-counting. This rectification process works as follows:

All n-grams will be ordered by number of words (n) first and their frequency (i.e., number of locations) within the collection of texts second. Each n-gram will then be evaluated in that order. If its frequency is high enough, it will be designated as a principal phrase. At that point, we remove the n-gram locations of each remaining n-gram that starts and/or ends within any of the locations of this designated principal phrase. This will reduce the frequency by 1 of each n-gram that has one of its locations removed.

**Algorithm:**

```
For each text:
   For each block:
      For each n-gram within the block:
         Starts with a stop word not equal to "not" or "no" – next
         Ends with a stop word – next
         n-gram appears in list of excluded phrases – next
         Store in list of n-gram locations with text, block, and position
         identifiers
Order the unique n-grams in the list of n-gram locations by n, highest first, then
by their overall frequency, highest first. Run through the n-grams in that order.
Rectification - For each n-gram:
   If current frequency greater than min. frequency:
      Designate as a principal phrase
```

```
Remove each n-gram location that appears in the same text and block of any of
the locations of the designated principal phrase, and for which its position
is such that it starts or ends within the positions of the applicable
location of the designated principal phrase. For each such an n-gram location
removed, reduce the frequency of that n-gram by 1.
```

The end result will be a selection of principal phrases with their locations, which we can use to determine the frequencies of each principal phrase in each text.

## 2.1. Example

Let's consider the sentence "The authors wrote abstract phrase mining papers." as part of a hypothetical situation where it occurs in a set of one or more larger texts. The question is, what are the principal phrases in this sentence, when we are looking for principal phrases that consist of at least 2 words? First, we find all the n-grams with n greater than or equal to 2. Since we will not have our n-grams cross punctuation marks, n is at most 7. However, "the" is a stop word, and thus no n-gram can start with it. As such the largest n-gram has n=6, and the 6-gram is "authors wrote abstract phrase mining papers". Table 1 shows all the n-grams we can find in this sentence.

*Table 1: n-Grams in " The authors wrote abstract phrase mining papers."*

| 2-grams | authors wrote<br>wrote abstract<br>abstract phrase<br>phrase mining<br>mining papers |
|---|---|
| 3-grams | authors wrote abstract<br>wrote abstract phrase<br>abstract phrase mining<br>phrase mining papers |
| 4-grams | authors wrote abstract phrase<br>wrote abstract phrase mining<br>abstract phrase mining papers |

| 5-grams | authors wrote abstract phrase mining |
| | wrote abstract phrase mining papers |
| 6-grams | authors wrote abstract phrase mining papers |

Each of these n-grams will be assigned a frequency of 1 due to our sentence.

We see that this one short sentence gives rise to 15 n-grams, most of which do not appear to be meaningful phrases. Which ones will be selected as principal phrases depends on the frequency of the same n-grams in other parts of the text(s).

Let's assume that the minimum frequency for any n-gram to be considered a principal phrase is 3, and that the only n-grams that made the frequency requirement in the sentence "The authors wrote abstract phrase mining papers" are the ones shown in Table 2.

*Table 2: Frequencies of n-grams in Hypothetical Texts*

| n-grams | frequency |
| --- | --- |
| abstract phrase | 10 |
| phrase mining | 18 |
| mining papers | 5 |
| authors wrote abstract | 3 |
| abstract phrase mining | 10 |
| phrase mining papers | 5 |

Note that the frequency of these phrases are the frequencies collected from the entire collection of texts under consideration.

The largest n here is 3, so we look for the 3-gram with the highest frequency. That is the phrase "abstract phrase mining". We designate this as a principal phrase.

Then we look for any n-grams whose words started or ended in one of the occurrences of this phrase. The n-gram "abstract phrase" occurs in each, so its frequency is reduced by 10. This causes it to be eliminated; it is NOT a principal phrase.

The n-gram "phrase mining" also wholly occurs in "abstract phrase mining", so this n-gram also loses 10 frequencies. In this case, however, 8 more frequencies are left.

In our sentence, "mining papers" starts in our principal phrase, so it would lose one frequency for that occurrence. There may be additional such occurrences in the remaining blocks/texts, let's say there are 3 of those. So "mining papers" ends up with a frequency below 3 and would be eliminated.

Similarly, "authors wrote abstract" ended inside the principal phrase "abstract phrase mining" for our sentence. As such it loses at least one frequency and is eliminated.

Finally, "phrase mining papers" also starts in our principal phrase in our sentence, so that one also loses at least one frequency. If there are more such cases, it may also end up below 3 and be eliminated.

In the end, all that is left besides "abstract phrase mining", is "phrase mining" with a frequency of 8. If its frequencies stay above 3 when all other principal phrases consisting of 3 words, and all principal phrases consisting of 2 words with a greater frequency, have been processed, it will also be designated as a principal phrase.

Note that the only two n-grams left from our sentence are meaningful ones. Also note that "phrase mining" ended up with a lower frequency than its super phrase "abstract phrase mining".

## 3. Results

As a real-world example, on March 31, 2022 we selected all publications in PubMed related to the Omicron variant of the Coronavirus. Using R, we imported the data into a corpus, then extracted all principal phrases with a minimum of two and a maximum of 8 words from their abstracts. The size of the file obtained from PubMed was 6207 KB and took a little over 1 second to read into a data table in R. Using this data table, we extracted 1375 principal phrases from the abstracts of 1531 documents in 16 seconds. The 10 most common phrases with their frequencies within this collection of texts are shown in Table 3.

*Table 3: Most Common Principal Phrases in the Omicron Dataset*

| *Principal Phrase* | *Frequency* |
| --- | --- |
| omicron variant | 451 |
| 95% ci | 153 |
| delta variant | 132 |
| spike protein | 131 |
| sars-cov-2 variants | 117 |
| variants of concern | 105 |
| severe acute respiratory syndrome coronavirus | 94 |
| sars-cov-2 infection | 91 |
| sars-cov-2 omicron variant | 87 |
| neutralizing antibodies | 83 |

Upon inspection of the first abstract in the collection, we found the principal phrases and frequencies as shown in Table 4.

*Table 4: Principal Phrases in the First Document of the Omicron Dataset*

| Principal Phrases for PMID: 34982466 | Frequency |
|---|---|
| amino acid | 1 |
| covid-19 vaccines | 1 |
| mutations in the spike protein | 2 |
| new variant | 1 |
| new variant of sars-cov-2 | 1 |
| november 2021 | 1 |
| omicron variant | 2 |
| previous variants | 1 |
| previously infected | 1 |
| receptor-binding domain | 2 |
| sars-cov-2 variants | 2 |
| south africa | 1 |
| spike protein | 1 |
| transmissibility and immune | 1 |
| vaccinated individuals | 1 |
| variant of concern | 1 |
| world health organization | 1 |

It is clear that the principal phrases are mostly meaningful. In the case that certain of the phrases are not considered meaningful, such as, for example, the phrase "transmissibility and immune", they can be excluded from the process, either at the start when the principal phrases are extracted, or afterwards. (Small, Cabrera, & Kostis, 2020) describe an application that extracts principal phrases from PubMed files and allows phrases that are not meaningful to the user to be removed. The application is created by the authors of this paper using the functionality described here, and implemented via functions in the `phm` R package.

## 4. Conclusion

**Strengths:**

a) We envisage a wealth of applications for this method; any application that currently extracts words from texts can be improved by selecting principal phrases instead; the information provided that way will be more specific and useful. An example of an application of our method is to extract principal phrases from a collection of text boxes from a dataset of medical records to generate features for outcomes prediction.

b) Sentiments of words that are preceded by "not" or "no" are no longer incorrectly identified.

c) In comparison to currently existing methods for extracting phrases from texts, there is no need to provide a list of quality phrases. As such, it is not possible for any important frequent phrases to be missed, which, using any of the existing methods, may be the case for new phrases, or phrases in a different language, or any other phrases not present on the provided lists of quality phrases.

**Limitations:**

a) Unlike existing methods for extracting phrases, this method may on occasion return a phrase that is not particularly meaningful. However, it is possible to remove such meaningless phrases once they have been identified.

b) When words are extracted from texts, we are able to use several lexicons to determine the sentiment of those words, and so we are able to determine the sentiment of each text as well as the overall sentiment of the complete collection. Since there are no such lexicons for phrases, this would be a disadvantage of our method. However, it may be possible in the future to establish a lexicon with sentiments for phrases.

c) Certain phrases may be very similar, such as, for example, "previous variant" and "previous variants", and it could be argued that they should be combined. This issue can be resolved by stemming the words in the texts first before extracting principal phrases, but it could also be a future enhancement to combine like phrases as part of the extraction process.

In conclusion, we have developed a method that will extract principal phrases from a collection of texts or a corpus of documents with texts, where a principal phrase is a phrase that occurs in this collection of texts, does not cross punctuation marks, does not start with a stop word with the exception of the stop words "not" and "no", does not end with a stop word, is frequent within those texts without having been double counted, and is meaningful to the user.

The method is implemented in the R-language via a package called `phm`, where the function `phraseDoc` will extract the principal phrases. This package also contains several other functions to assist in obtaining information from those texts utilizing the principal phrases. A vignette called `phm-intro` has been included to explain the process and functions of this package.

Note also that the dataset and R code used for the example in this paper are available for perusal.